\documentclass[journal]{IEEEtran}

\usepackage{multirow}
\usepackage{cite}
\usepackage{amsmath,amssymb,amsfonts,nicefrac}
\usepackage{algorithmic}
\usepackage{graphicx}
\usepackage{textcomp}
\usepackage{xcolor}

\ifCLASSINFOpdf
\else
\fi

\begin{document}

\title{Feature-level Rating System using Customer Reviews and Review Votes}
\author{Koteswar~Rao~Jerripothula,~\IEEEmembership{Member,~IEEE,}
        Ankit~Rai, Kanu~Garg, 
        and Yashvardhan~Singh~Rautela
\thanks{
Koteswar Rao Jerripothula is with the CSE Department, Indraprastha Institute of Information Technology Delhi (IIIT Delhi), India. The remaining authors are with the CSE Department, Graphic Era, India. Contact Author: Koteswar Rao Jerripothula. Email: koteswar@iiitd.ac.in}

}

\markboth{Accepted by IEEE Transactions on Computational Social Systems}%
{Jerripothula \MakeLowercase{\textit{et al.}}: Feature-level Rating System using Customer Reviews and Review Votes}

\maketitle

\begin{abstract}
This work studies how we can obtain feature-level ratings of the mobile products from the customer reviews and review votes to influence decision making, both for new customers and manufacturers. Such a rating system gives a more comprehensive picture of the product than what a product-level rating system offers. While product-level ratings are too generic, feature-level ratings are particular; we exactly know what is good or bad about the product. There has always been a need to know which features fall short or are doing well according to the customer’s perception. It keeps both the manufacturer and the customer well-informed in the decisions to make in improving the product and buying, respectively. Different customers are interested in different features. Thus, feature-level ratings can make buying decisions personalized. We analyze the customer reviews collected on an online shopping site (Amazon) about various mobile products and the review votes. Explicitly, we carry out a feature-focused sentiment analysis for this purpose. Eventually, our analysis yields ratings to 108 features for 4k+ mobiles sold online. It helps in decision making on how to improve the product (from the manufacturer’s perspective) and in making the personalized buying decisions (from the buyer’s perspective) a possibility. Our analysis has applications in recommender systems, consumer research, etc.
\end{abstract}


\IEEEpeerreviewmaketitle

\section{Introduction}
With the rise of the internet and the kind of busy lifestyles people have today, online shopping has become a norm. Customers often rely on the online ratings of the previous customers to make their decisions. However, most of these ratings on the online websites are product-level ratings and lack specificity. Although products can be compared based on the product-level ratings available, there is always a class of people who prefer buying the items based on particular features. Such people have to generally go through the entire comments section to know previous customers’ perceptions \cite{raghavan2014modeling,farasat2016social} of the product’s features in which they are interested. Considering the number of products present for an item (such as mobile), it becomes a tedious job for a customer to arrive at the best product for himself. Moreover, from a manufacturer’s perspective, such product level ratings hardly specify what is good or bad about the product \cite{zhu2017maximizing}. So, if feature level ratings are available, it gives more clarity to the seller on how to improve the product. Given all these benefits, our goal is to develop a feature-level rating system.   

\begin{figure}
	\centering
	\includegraphics[width=0.95\linewidth]{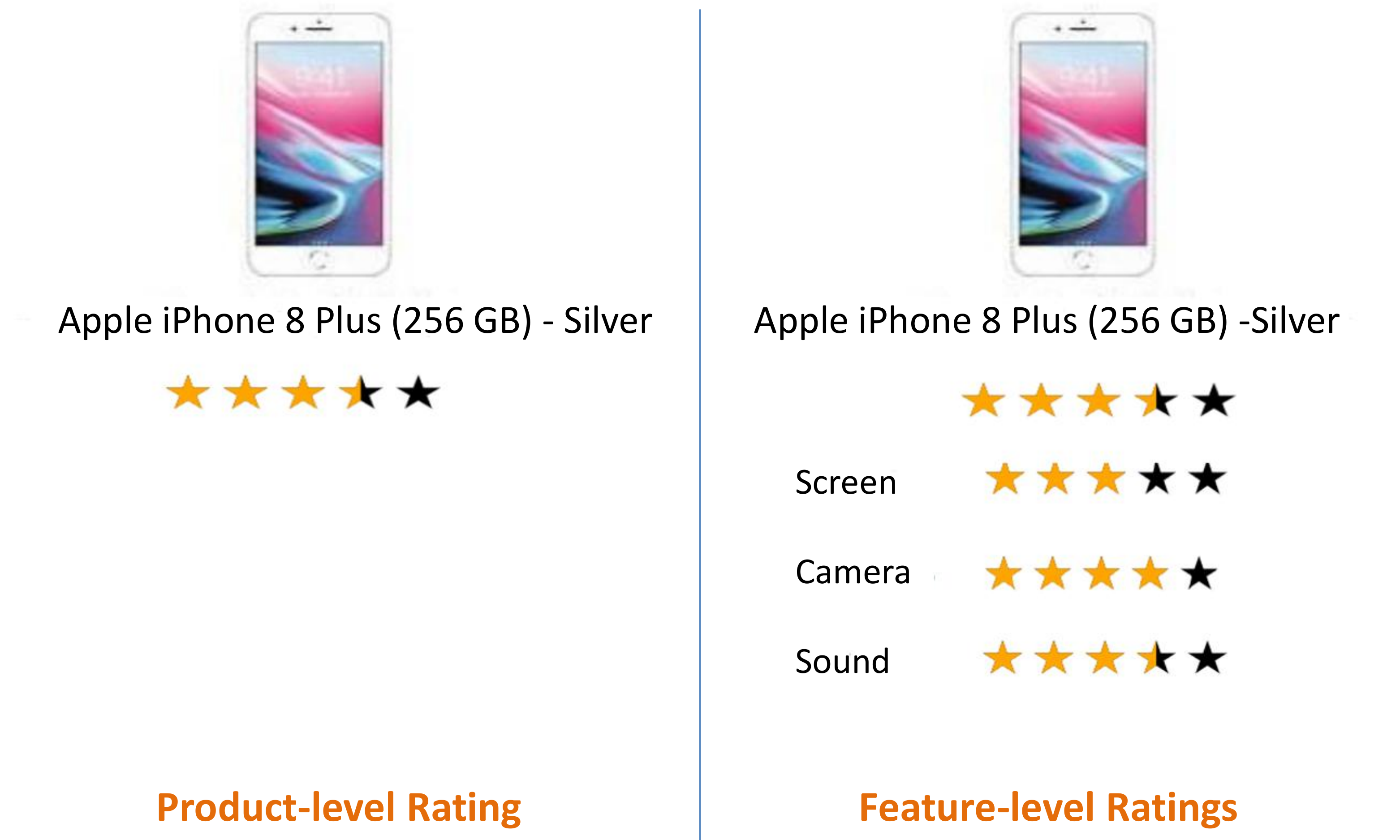}
	\caption{Product-level Rating System versus Feature-level Rating System: While the first one is very generic, the latter is quite specific.}\label{fig:pvsf}
	\centering
\end{figure}

Although feature-level ratings can also be requested from the customers just like the product-level ratings, it is not a good proposal, for there could be too many features. Instead, it is much more practical to leverage whatever reviews and review votes \cite{yang2017collaborative,karan2016dynamics} that are being already given by customers to provide feature-level ratings. The reviews are made up of sentences, and every sentence has some sentiment\cite{smarandache2019word,ravi2017fuzzy,xu2019sentiment,khan2016sentimi} associated with it, viz. positive, neutral, or negative. Also, since they can be separated, we can always extract the sentences describing a particular feature of the product and subsequently obtain sentiment scores over such sentences. By utilizing these sentiment scores as the basis and the review votes as a support, we can build a feature-level rating system that can yield feature-level ratings, as shown in Fig.~\ref{fig:pvsf}.

However, there are few challenges in building such a feature-level rating system. First, we need to determine which features to look for in an item. Another challenge is that there could be many words relating to the same feature; they all need to be clubbed into one feature. Second, we have to pre-process the data as some of the review comments may contain non-English languages, one word, spelling mistakes, etc. Third, we have to devise a way to transform the extracted sentiment scores into an appropriate rating for a feature of a product while incorporating the review votes.

As far as feature identification of an item (such as mobile) is concerned, we go through the word frequency table of the entire customer review data for that item, at least up to a particular frequency. Next, all the related words coming under the same feature are grouped, and the most frequent one is chosen as a representative.  We call such representatives as feature keywords. Then, we perform a series of pre-processing steps to filter out the unnecessary data, correct the remaining, and turn it into structured data. Each review is broken into sentences, and only relevant sentences are retained. The relevant sentences are passed through sentiment analyzer to generate sentiment scores, which are then adjusted to the ratings. Scores within a particular range are given a specific rating. The ratings of the relevant sentences containing a particular feature are combined using the weighted-average \cite{8269367} to obtain the final ratings since all opinions are not equally valuable. We leverage review-votes to assign the required weights.

Our contributions are as follows:
We develop a feature-level rating system that takes customer reviews and review votes as inputs and outputs feature-level ratings.
We obtain such ratings for as many as 4k+ mobiles sold online in terms of as many as 108 features.
We propose votes-aware cumulative rating and votes-aware final rating measures, a new way of accumulating and finalizing the sentiment scores.
Although there are no ground truths available, we still manage to evaluate our approach by comparing the final ratings of our phone feature against overall ratings of the phone given by the customers themselves, which leads to remarkable results demonstrating the effectiveness of our method.   

\section{Related Work}

Sentiment analysis \cite{wiliam2020sentiment,singh2020she,lin2018leveraging,bui2016temporal,chen2019user,kumar2020movie,ling2020hybrid,chakraborty2020survey} has been an active research topic for a long period now. It has applications in health \cite{yang2016mining,palomino2016online,khan2016sentiment}, politics \cite{ramteke2016election,proksch2019multilingual}, sports \cite{yu2015world,lucas2017goaalll}, e-commerce \cite{addepalli2016proposed,mehta2016sentiment}, etc. In e-commerce, customer reviews can give lots of insights about the products, as shown in \cite{Fang2015,raja2017feature}, through sentiment analysis. Specifically, \cite{wiliam2020sentiment} studies trends of mobile brands on Twitter through sentiment analysis. However, the analysis is restricted to the mobile overall, not to specific features. \cite{nandal2020machine} tries to do so, but for limited products and limited features. While \cite{nandal2020machine} used SVM, a supervised learning algorithm, \cite{sadhasivam2019sentiment} used ensembling for achieving this. In this paper, we attempt to exploit these customer reviews to provide ratings for as many as 108 features of 4k+ mobile phones sold online while incorporating review votes, which has never been done in the previous studies. Moreover, we do this in an unsupervised way, not supervised or weakly-supervised \cite{8290832,8099896,10.1007/978-3-319-46478-7_12,7025663} way, thanks to the lexical approach of generating sentiment scores for a sentence. Similar to our work, \cite{inproceedings} explored digital cameras and television for the same. However, they explore only ten features. \cite{bafna2013feature} explored only Cannon Camera, iPhone 4s, and Mp3 player. However, we explore as many as 4k+ products and provide our recommendations. Nevertheless, to the best of our knowledge, this is the first paper to account for review votes in a feature-level rating system.

\section{Methodology}
The proposed method has four steps: (1) feature selection; (2) pre-processing; (3) relevant sentence extraction; (4) feature-based rating generation. Every step is described in detail in this section.

\subsection{Feature Selection}
Let us say we collect a dataset of $N$ feature-related words, denoted by $\mathcal{W}=\{w_1,\cdots,w_N\}$, by manually going through word frequency table of the entire customer review data on an item (mobile, in our case). In this way, we identify the features in which people are generally interested. Note that we neglect the words having their frequency less than 0.02\% of the total number of reviews in the review data, which means they are rarely discussed feature-related words and can, therefore, be neglected. Let us say the corresponding frequencies of the feature-related words form another set denoted as $\mathcal{Z}=\{z_1,\cdots,z_N\}$. Since the feature-related words related to a particular feature should be clubbed into one feature, we define a relationizer function denoted as $\mathcal{R}(\mathcal{W},w_i)$, which returns a set of all the related words of $w_i$ in $\mathcal{W}$, including itself. Note that the relationizer function discussed here as a matter of notation is manual. We now define our feature dataset, denoted by $\mathcal{F}$, as a set of such distinct sets of related feature words, as defined below:
\begin{equation}
\mathcal{F}=\{\mathcal{R}(\mathcal{W},w_i)\}_{i\in\{1,\cdots,N\}},
\end{equation} where we iterate through all the words in $\mathcal{W}$ and form distinct sets of the related words using the relationizer function. Since different related words will form the same sets, the duplicates will be removed to make the sets left distinct. Now, let $\mathcal{F}_k$ be the $k^{th}$ feature words set in the $\mathcal{F}$ feature dataset.     
In any $\mathcal{F}_k$, a representative feature word is selected to identify the whole feature words set. Let us call such representatives as feature keywords. The most frequent feature word in the set is chosen as the representative (inspired by \cite{7351686}) or feature keyword to assign its name to the set, as shown below:
\begin{equation}
\mathcal{F}_k \leftarrow w_i|i=max(\{\mathcal{Z}(i)|w_i\in\mathcal{F}_k\}),  
\end{equation} where the feature set is assigned a name with the word that has the maximum frequency in the set. From now on, abusing notations a bit, $\mathcal{F}_k$ can mean both the $k^{th}$ feature set (a set of related feature words) and its keyword (or set's name), according to the context. 
\begin{table}
	\centering
	\caption{We remove all other characters except these, for they have some purpose or other in describing a feature in a sentence.}\label{tab:cr}
	\begin{center}
		\begin{tabular}{|l|l|}
			\hline
			Purpose & Characters\\\hline
			Word Formation& A-Za-z\\
			Punctuation & .,:;-!? (space)\\
			Emoticons & ':-()=*83$><$\^{}/[]\#\{\}$|$;\textbackslash\&\\
			\hline            
		\end{tabular}
	\end{center}
	\centering
\end{table}

\begin{figure}
	\centering
	\includegraphics[width=0.95\linewidth]{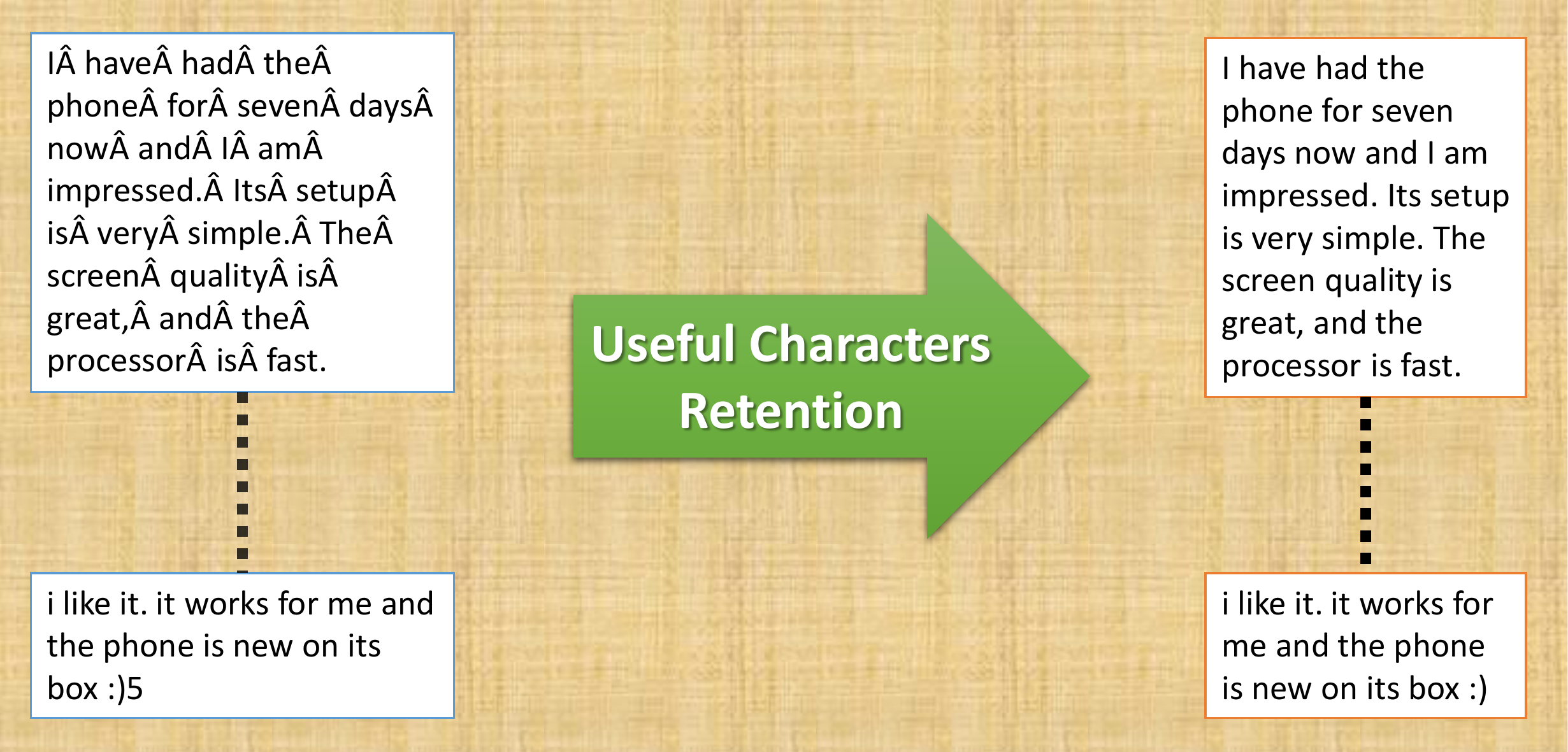}
	\caption{Pre-processing: Useful Characters Retention, where unnecessary characters are removed in this illustration.}\label{fig:useful}
	\centering
\end{figure}

\subsection{Pre-processing}
The review comments data is generally unstructured, for it is written by the customers online. Our goal now is to convert this unstructured data into structured data in our pre-processing steps, which means useful data is extracted, disintegrated, and corrected. 

Note that the data retained after removing the characters that are useless for our purpose is what we mean as useful data. While inspecting the reviews for figuring out the features to work with, we also noticed how people praise or criticize. People often use the characters required for adjective words, punctuation, or emoticons. While we retain the characters required for word formation, punctuation, and emoticons, we remove all other characters, including numbers, as shown in Fig.~\ref{fig:useful}. In Table.~\ref{tab:cr}, we give the information regarding what all characters are retained. After that, we remove any entries which are left empty because we have no use of them in the feature-level rating system. Let us consider that, for a product (not an item), we denote product review data as $\mathcal{D}=\{C_1,C_2,\cdots,C_m\}$, comprising of $m$ useful review comments. Note that when we said customer review data of the item in the last section for feature selection, we meant review data of all the mobiles. In contrast, $\mathcal{D}$ is the review data of just the mobile product under consideration.    

\begin{figure}
	\centering
	\includegraphics[width=0.95\linewidth]{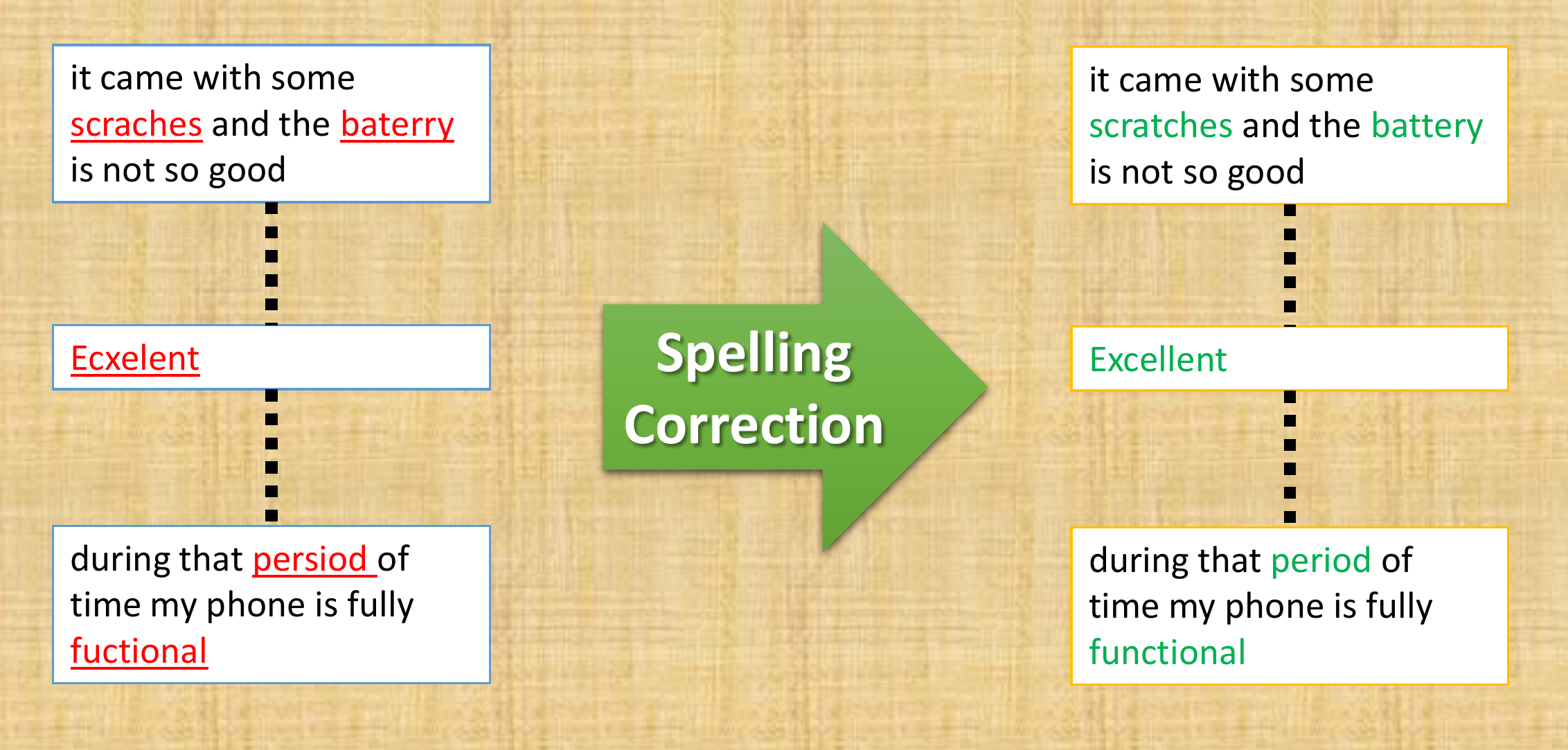}
	\caption{Pre-processing: Spelling Correction, where misspelled words like `scraches', `baterry', `Ecxelent', `persiod' and `fuctional' are corrected in this illustration.}\label{fig:spelling}
	\centering
\end{figure}

\begin{figure}
	\centering
	\includegraphics[width=0.95\linewidth]{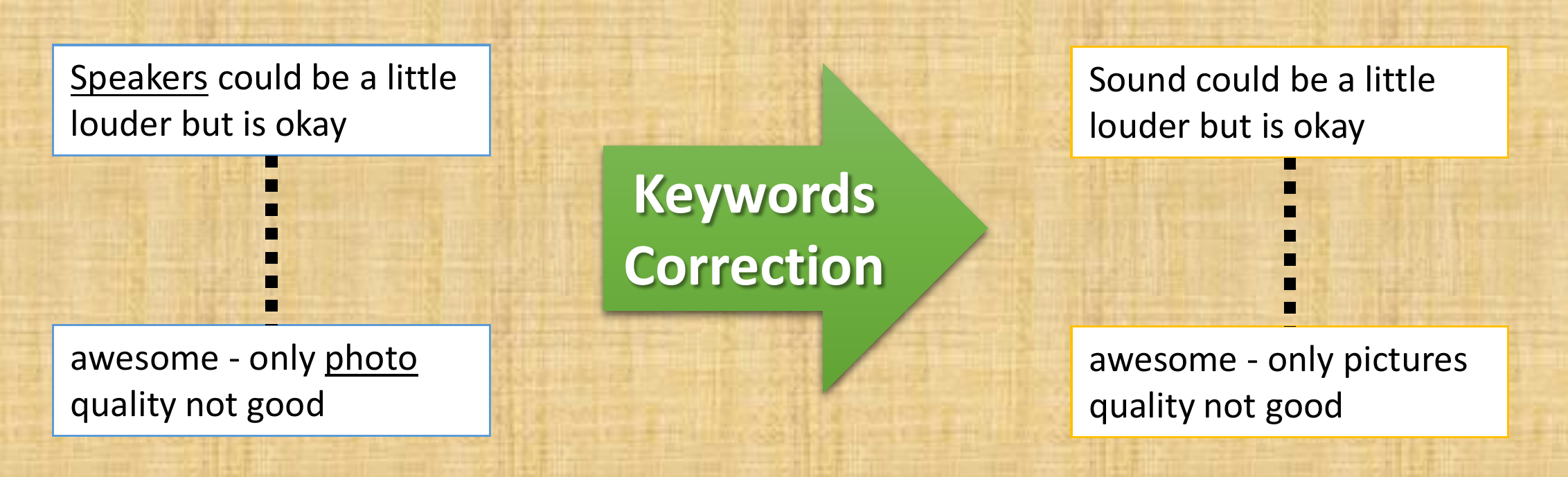}
	\caption{Pre-processing: Keyword Correction, where feature words like `speakers' and `photo' are corrected to their respective feature keywords, `sound' and `pictures' in this illustration.}\label{fig:synonyms}
	\centering
\end{figure}

By corrected data, we mean the data obtained after correcting the related feature words issue and spelling correction in the useful data just extracted. To correct the data in such a manner, we need to disintegrate the reviews into words and process them separately. We use the NLTK package of python for this purpose. It helps in disintegrating the reviews into words as tokens while neglecting the spaces. It considers even period (.) as a token, which becomes useful later while breaking the comments into sentences. Each comment can now be represented as a set of tokens, i.e. $C_j=\{t^1_j,t^2_j,\cdots,t^{|C_j|}_j\}$, where $|C_j|$ denotes the number of tokens obtained in $C_j$ and $t^i_j$ represents $i^{th}$ token of $j^{th}$ comment. We correct any token $t^i_j$ of the useful data in the following manner:

\begin{equation}\label{ceq}
t^i_j= 
\begin{cases}
\mathcal{F}_k,& \text{if }S(t^i_j)\in \mathcal{F}_k \text{ or } t_i\in \mathcal{F}_k, \forall \mathcal{F}_k \in \mathcal{F}\\
S(t^i_j),              & \text{otherwise,}
\end{cases}
\end{equation}
where $S(\cdot)$ represents spelling correcting function (using autocorrect package of python). If a token before or after spelling correction matches with any of the members in any of our feature words sets, we replace it with the feature keyword of that set; otherwise, we replace it with the corrected token. In this way, we take care of both the related words issue (by replacing with keywords) and the spelling correction issue simultaneously. The illustrations of spelling and keywords correction are given in Fig.~\ref{fig:spelling} and Fig.~\ref{fig:synonyms}. Thus, with the useful data extracted, disintegrated, and corrected, our reviews data for a product becomes structured. Now, we can say that $t^i_j$ is $i^{th}$ token of $j^{th}$ comment of $\mathcal{D}$.            

\begin{figure}
	\includegraphics[width=0.95\linewidth]{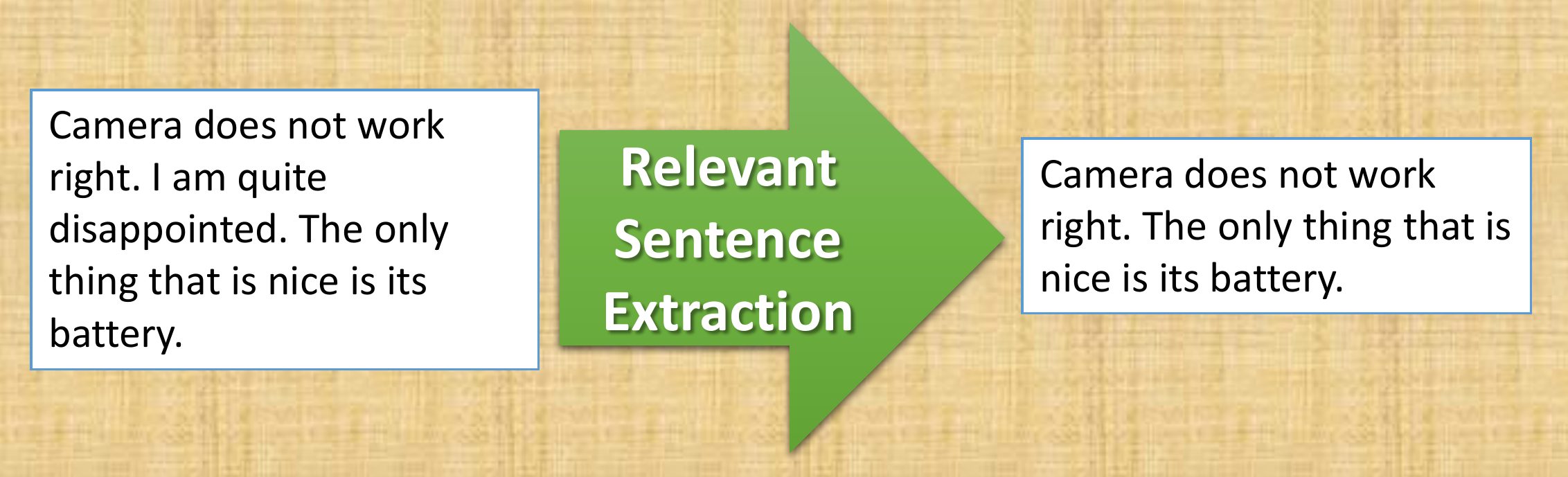}
	\caption{Relevant Sentence Extraction: Only sentences containing feature keywords are retained.}\label{fig:sentences}
\end{figure}

\begin{figure*}
	\includegraphics[width=0.95\linewidth]{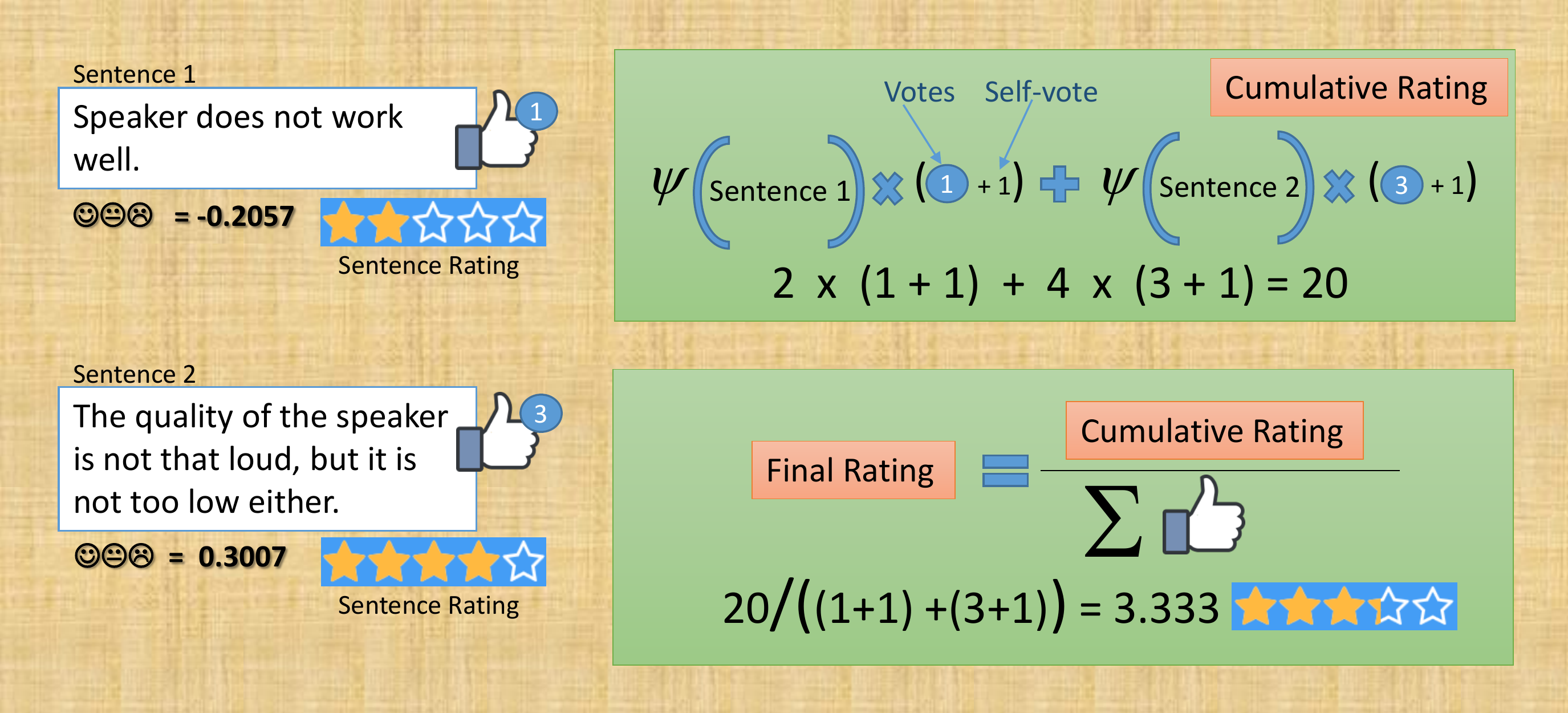}
	\caption{First, to compute cumulative rating for a feature, we accumulate sentence rating of all the sentences having the feature word (speaker in this case) along with their total votes (other + self). Second, to compute the final rating, we divide the cumulative rating by the total number of votes.}\label{fig:scoring}
\end{figure*}

\subsection{Relevant Sentence Extraction}
Since customer review data is now structured, we can group the continuous tokens present in a review as a sentence, as shown below in the definition of a set of sentences $X_j$ derived from $C_j$:

\begin{equation}
\begin{split}
X_j=\{(t^u_j,\cdots,t^v_j)|(t^v_j,t^{u-1}_j)=\text{`.'}\text{ , }(t^u_j,\cdots,t^{v-1}_j)\neq \text{`.'}, \\ (u,v)\in \{1,\cdots,|C_j|\} \text{ and } v>u\}
\end{split}
\end{equation}
where we call a group of continuous tokens as a sentence if the last and previous-to-beginning token are periods (.), and if all other tokens in that group are not periods (.). However, not all the sentences are relevant for feature-based rating. We define if a sentence $X_j^l$, $l^{th}$ sentence in $X_j$, is relevant or not in the following way:  
\begin{equation}
\rho(X_j^l) =
\begin{cases}
1, \text{ if } \mathcal{F}_k\in X_j^l \text{ for any } \mathcal{F}_k \in \mathcal{F}\\
0, \text{ otherwise}
\end{cases}
\end{equation}
where $\rho(\cdot)$ is relevance function for a sentence which outputs 1 if any of our feature keywords are present in the sentence. In this way, we extract only relevant sentences. See Fig.~\ref{fig:sentences} for an example.

\subsection{Feature-based Ratings Generation}

Having extracted relevant sentences, we can go through each sentence to figure out if it mentions a particular feature, say $F_k$. If yes, we can extract the emotion of the sentence to score it. For this purpose, we extract sentiment analysis scores \cite{hutto2014vader} for each of these sentences. We use \cite{hutto2014vader} because it accommodates emoticons also while performing the analysis. We use their compound score as the required sentiment analysis score. It ranges between -1 and 1. We divide this range into five equal parts and assign the ratings progressively, as shown in the Table~\ref{tab:stars}. Let the function that computes the sentiment score and assigns the appropriate rating be $\psi(\cdot)$. Then, we compute cumulative rating ($\mathcal{Q}(\cdot)$) for each feature over the entire product review data, i.e., $\mathcal{D}$, in the following manner: 

\begin{equation}
\mathcal{Q}(\mathcal{F}_k)=\sum\limits_{C_j\in\mathcal{D}}\sum\limits_{X^l_j \in X_j}^{\rho({X^l_j})=1} \psi(X^l_j)\times \delta(\mathcal{F}_k\in X^l_j)\times(\phi(C_j)+1)
\end{equation}  
where $\delta(\cdot)$ denotes logical function to check if a sentence consists of the concerned feature or not. During our accumulation, we also consider the number of votes received to review to which the sentence belongs. These votes inform us about the strength of the opinion associated with the reviews. Let $\phi(C_j)$ denote the number of votes received for $C_j$. Here, we are assuming that any sentence equally contributes to the strength of the opinion. We adjust the votes by adding 1 to account for self-votes of the customers who originally wrote the reviews. This accumulation is illustrated in Fig.~\ref{fig:scoring}. Then, we compute our final rating ($\mathcal{A}(\cdot)$) for a feature $\mathcal{F}_k$ using the below equation:  
\begin{equation}
\mathcal{A}(\mathcal{F}_k)=\frac{\mathcal{Q}(\mathcal{F}_k)}{\sum\limits_{C_j\in\mathcal{D}}\sum\limits_{X^l_j \in X_j}^{\rho({X^l_j})=1} \delta(\mathcal{F}_k\in X^l_j)\times(\phi(C_j)+1)}
\end{equation}
where we divide the cumulative number of stars by the total number of votes received during the accumulation. In this way, we are essentially computing the weighted average \cite{7457899}, where weights are determined by the votes received. So, we now have a feature-level rating for the feature $\mathcal{F}_k$ of a product using the customer reviews and review votes. The same proposed methodology can be applied to any number of features, any number of products, and any number of items. 

\begin{table}
	\begin{center}
		\caption{Conversion of a sentence's sentiment score to sentence's rating}\label{tab:stars}
		\begin{tabular}{|c|c|l|}
			\hline
			Sentiment Score & Rating & Meaning
			
			\\\hline
			-1.0 to -0.6 & 1-star & Terrible\\\hline
			-0.6 to -0.2 & 2-star& Poor  \\\hline
			-0.2 to 0.2 & 3-star & Average\\\hline
			0.2 to 0.6 & 4-star & Very Good\\\hline
			0.6 to 1.0 & 5-star & Excellent          \\\hline
		\end{tabular}
	\end{center}
\end{table}

\begin{table*}
	\caption{Our features dataset: the related words are separated by $\textcolor{red}{||}$ and led by a \textcolor{blue}{\textbf{keyword}} }\label{tab:words}
	\begin{center}
		\begin{tabular}{|l|}
			\hline
			$\text{ }\boldsymbol{\textcolor{red}{||}}\text{ }$\textcolor{blue}{\textbf{phone}},    product,    phones,    device,    cell,    item,    smartphone,    mobile,    model,    cellphone,    products,    devices,    piece,    cellular,    smartphones,    items,    telephone,\\\hline    handset,cellphones$\text{ }\boldsymbol{\textcolor{red}{||}}\text{ }$ \textcolor{blue}{\textbf{screen}},    display,    glass,    screens,    lcd,    displays,    displaying$\text{ }\boldsymbol{\textcolor{red}{||}}\text{ }$\textcolor{blue}{\textbf{battery}},    batteries$\text{ }\boldsymbol{\textcolor{red}{||}}\text{ }$\textcolor{blue}{\textbf{camera}},    resolution,    cameras,    pixels,    pixel,    cam,\\\hline    megapixel,      megapixels$\text{ }\boldsymbol{\textcolor{red}{||}}\text{ }$\textcolor{blue}{\textbf{price}},    money,    buying,    pay,    cost,    paid,    sold,    purchasing,    spent,    budget,    bucks,    rate,    spending,    pricing,    pays$\text{ }\boldsymbol{\textcolor{red}{||}}\text{ }$\textcolor{blue}{\textbf{sim}},    card,    dual,\\\hline    sims$\text{ }\boldsymbol{\textcolor{red}{||}}\text{ }$\textcolor{blue}{\textbf{apps}},     app,    program,    application,    widgets,    processes,    module$\text{ }\boldsymbol{\textcolor{red}{||}}\text{ }$\textcolor{blue}{\textbf{android}},    version,    operating,    ios,    versions,    software,    windows$\text{ }\boldsymbol{\textcolor{red}{||}}\text{ }$\textcolor{blue}{\textbf{case}},    box,\\\hline    packaged,    packing,     boxes,    packs$\text{ }\boldsymbol{\textcolor{red}{||}}\text{ }$\textcolor{blue}{\textbf{charge}},    charged,    charging,    charges,    discharges,    discharge,    charging$\text{ }\boldsymbol{\textcolor{red}{||}}\text{ }$\textcolor{blue}{\textbf{charger}},    plug,    adapter,    chargers,    plugs\\\hline$\text{ }\boldsymbol{\textcolor{red}{||}}\text{ }$\textcolor{blue}{\textbf{service}},    services$\text{ }\boldsymbol{\textcolor{red}{||}}\text{ }$\textcolor{blue}{\textbf{watch}},      clock$\text{ }\boldsymbol{\textcolor{red}{||}}\text{ }$\textcolor{blue}{\textbf{size}},    sizes$\text{ }\boldsymbol{\textcolor{red}{||}}\text{ }$\textcolor{blue}{\textbf{call}},    talk, voice, called, talking,    dial,    speak,    outgoing,    communications,    communicating$\text{ }\boldsymbol{\textcolor{red}{||}}\text{ }$\textcolor{blue}{\textbf{wifi}}\\\hline$\text{ }\boldsymbol{\textcolor{red}{||}}\text{ }$\textcolor{blue}{\textbf{brand}}$\text{ }\boldsymbol{\textcolor{red}{||}}\text{ }$\textcolor{blue}{\textbf{memory}},    data,    space  $\text{ }\boldsymbol{\textcolor{red}{||}}\text{ }$\textcolor{blue}{\textbf{pictures}},    picture,    photos,    pics,    photo,    image,    images,    photography$\text{ }\boldsymbol{\textcolor{red}{||}}\text{ }$\textcolor{blue}{\textbf{touch}},    touchscreen$\text{ }\boldsymbol{\textcolor{red}{||}}\text{ }$\textcolor{blue}{\textbf{text}},    texts,    editing, \\\hline    txt,    texted$\text{ }\boldsymbol{\textcolor{red}{||}}\text{ }$\textcolor{blue}{\textbf{sound}},    speaker,      speakers,     sounds,    loudspeaker$\text{ }\boldsymbol{\textcolor{red}{||}}\text{ }$\textcolor{blue}{\textbf{sd}},    micro,    microsd$\text{ }\boldsymbol{\textcolor{red}{||}}\text{ }$\textcolor{blue}{\textbf{network}},    networks$\text{ }\boldsymbol{\textcolor{red}{||}}\text{ }$\textcolor{blue}{\textbf{button}},    keyboard,    buttons,    key,    keys, \\\hline    typing,    qwerty,    keypad,    keyboards,    dials$\text{ }\boldsymbol{\textcolor{red}{||}}\text{ }$\textcolor{blue}{\textbf{music}},     audio,    listen,    listening$\text{ }\boldsymbol{\textcolor{red}{||}}\text{ }$\textcolor{blue}{\textbf{internet}},    online,    web,    browsing$\text{ }\boldsymbol{\textcolor{red}{||}}\text{ }$\textcolor{blue}{\textbf{light}},    flash,    flashlight,    torch\\\hline$\text{ }\boldsymbol{\textcolor{red}{||}}\text{ }$\textcolor{blue}{\textbf{warranty}}$\text{ }\boldsymbol{\textcolor{red}{||}}\text{ }$\textcolor{blue}{\textbf{bluetooth}},    wireless$\text{ }\boldsymbol{\textcolor{red}{||}}\text{ }$\textcolor{blue}{\textbf{video}},     videos,    streaming,      stream,    fps$\text{ }\boldsymbol{\textcolor{red}{||}}\text{ }$\textcolor{blue}{\textbf{settings}},    setting,    setup,    configure,    configuration,    calibration$\text{ }\boldsymbol{\textcolor{red}{||}}\text{ }$\textcolor{blue}{\textbf{color}},    \\\hline colors,    colour$\text{ }\boldsymbol{\textcolor{red}{||}}\text{ }$\textcolor{blue}{\textbf{design}},    build,    shape,    compact$\text{ }\boldsymbol{\textcolor{red}{||}}\text{ }$\textcolor{blue}{\textbf{download}},      downloaded,    downloads$\text{ }\boldsymbol{\textcolor{red}{||}}\text{ }$\textcolor{blue}{\textbf{usb}},    microusb$\text{ }\boldsymbol{\textcolor{red}{||}}\text{ }$\textcolor{blue}{\textbf{email}},    emails$\text{ }\boldsymbol{\textcolor{red}{||}}\text{ }$\textcolor{blue}{\textbf{speed}},    speeds,    speedy\\\hline$\text{ }\boldsymbol{\textcolor{red}{||}}\text{ }$\textcolor{blue}{\textbf{headphones}},    headset,    earphones,    earphone,    headsets $\text{ }\boldsymbol{\textcolor{red}{||}}\text{ }$\textcolor{blue}{\textbf{gps}}$\text{ }\boldsymbol{\textcolor{red}{||}}\text{ }$\textcolor{blue}{\textbf{games}},     gaming$\text{ }\boldsymbol{\textcolor{red}{||}}\text{ }$\textcolor{blue}{\textbf{ram}}$\text{ }\boldsymbol{\textcolor{red}{||}}\text{ }$\textcolor{blue}{\textbf{messages}},    messaging,    sms,    messenger,    msg$\text{ }\boldsymbol{\textcolor{red}{||}}\text{ }$\textcolor{blue}{\textbf{cable}},\\\hline    cord,    connector,    cables$\text{ }\boldsymbol{\textcolor{red}{||}}\text{ }$\textcolor{blue}{\textbf{manual}},    instructions,    instruction,    booklet  $\text{ }\boldsymbol{\textcolor{red}{||}}\text{ }$\textcolor{blue}{\textbf{processor}},    cpu,      processors$\text{ }\boldsymbol{\textcolor{red}{||}}\text{ }$\textcolor{blue}{\textbf{specs}},    specifications,    spec$\text{ }\boldsymbol{\textcolor{red}{||}}\text{ }$\textcolor{blue}{\textbf{hardware}}\\\hline$\text{ }\boldsymbol{\textcolor{red}{||}}\text{ }$\textcolor{blue}{\textbf{fingerprint}},    finger,    fingers,    fingerprints$\text{ }\boldsymbol{\textcolor{red}{||}}\text{ }$\textcolor{blue}{\textbf{switch}},    switched,    switches$\text{ }\boldsymbol{\textcolor{red}{||}}\text{ }$\textcolor{blue}{\textbf{accessories}},    accessory$\text{ }\boldsymbol{\textcolor{red}{||}}\text{ }$\textcolor{blue}{\textbf{weight}},      bulky,    lightweight$\text{ }\boldsymbol{\textcolor{red}{||}}\text{ }$\textcolor{blue}{\textbf{sensor}},    sensors\\\hline$\text{ }\boldsymbol{\textcolor{red}{||}}\text{ }$\textcolor{blue}{\textbf{face}},    faces$\text{ }\boldsymbol{\textcolor{red}{||}}\text{ }$\textcolor{blue}{\textbf{protection}},    firmware,    protectors,    virus,    antivirus$\text{ }\boldsymbol{\textcolor{red}{||}}\text{ }$\textcolor{blue}{\textbf{notification}},    notifier, notifications, prompts$\text{ }\boldsymbol{\textcolor{red}{||}}\text{ }$\textcolor{blue}{\textbf{brightness}}$\text{ }\boldsymbol{\textcolor{red}{||}}\text{ }$\textcolor{blue}{\textbf{jack}} $\text{ }\boldsymbol{\textcolor{red}{||}}\text{ }$\textcolor{blue}{\textbf{chip}},\\\hline    chipset$\text{ }\boldsymbol{\textcolor{red}{||}}\text{ }$\textcolor{blue}{\textbf{tested}},    checked,    tests$\text{ }\boldsymbol{\textcolor{red}{||}}\text{ }$\textcolor{blue}{\textbf{scanner}},    qr$\text{ }\boldsymbol{\textcolor{red}{||}}\text{ }$\textcolor{blue}{\textbf{files}}$\text{ }\boldsymbol{\textcolor{red}{||}}\text{ }$\textcolor{blue}{\textbf{microphone}}$\text{ }\boldsymbol{\textcolor{red}{||}}\text{ }$\textcolor{blue}{\textbf{navigation}},    navigating$\text{ }\boldsymbol{\textcolor{red}{||}}\text{ }$\textcolor{blue}{\textbf{waterproof}}$\text{ }\boldsymbol{\textcolor{red}{||}}\text{ }$\textcolor{blue}{\textbf{calendar}}$\text{ }\boldsymbol{\textcolor{red}{||}}\text{ }$\textcolor{blue}{\textbf{mms}}$\text{ }\boldsymbol{\textcolor{red}{||}}\text{ }$ \textcolor{blue}{\textbf{alarm}},    alarms\\\hline$\text{ }\boldsymbol{\textcolor{red}{||}}\text{ }$\textcolor{blue}{\textbf{hotspot}}$\text{ }\boldsymbol{\textcolor{red}{||}}\text{ }$\textcolor{blue}{\textbf{graphics}},    gpu,    graphic$\text{ }\boldsymbol{\textcolor{red}{||}}\text{ }$\textcolor{blue}{\textbf{icons}},    icon$\text{ }\boldsymbol{\textcolor{red}{||}}\text{ }$\textcolor{blue}{\textbf{selfie}},    selfies$\text{ }\boldsymbol{\textcolor{red}{||}}\text{ }$\textcolor{blue}{\textbf{recharge}},    recharging$\text{ }\boldsymbol{\textcolor{red}{||}}\text{ }$\textcolor{blue}{\textbf{electronics}}$\text{ }\boldsymbol{\textcolor{red}{||}}\text{ }$\textcolor{blue}{\textbf{vibrate}},    vibration, vibrates,    vibrating,     \\\hline shake$\text{ }\boldsymbol{\textcolor{red}{||}}\text{ }$\textcolor{blue}{\textbf{recording}},    recorder$\text{ }\boldsymbol{\textcolor{red}{||}}\text{ }$\textcolor{blue}{\textbf{zoom}},    lens$\text{ }\boldsymbol{\textcolor{red}{||}}\text{ }$\textcolor{blue}{\textbf{chat}},    chatting$\text{ }\boldsymbol{\textcolor{red}{||}}\text{ }$\textcolor{blue}{\textbf{ringtone}},    ringing$\text{ }\boldsymbol{\textcolor{red}{||}}\text{ }$\textcolor{blue}{\textbf{heats}},    overheats,    temperature,    overheated,    temp$\text{ }\boldsymbol{\textcolor{red}{||}}\text{ }$\textcolor{blue}{\textbf{voicemail}},     \\\hline inbox$\text{ }\boldsymbol{\textcolor{red}{||}}\text{ }$\textcolor{blue}{\textbf{stereo}}  $\text{ }\boldsymbol{\textcolor{red}{||}}\text{ }$\textcolor{blue}{\textbf{scroll}},    slider,    slides,    swipe$\text{ }\boldsymbol{\textcolor{red}{||}}\text{ }$\textcolor{blue}{\textbf{guarantee}},    guaranteed$\text{ }\boldsymbol{\textcolor{red}{||}}\text{ }$\textcolor{blue}{\textbf{languages}}$\text{ }\boldsymbol{\textcolor{red}{||}}\text{ }$\textcolor{blue}{\textbf{multimedia}}$\text{ }\boldsymbol{\textcolor{red}{||}}\text{ }$\textcolor{blue}{\textbf{compatibility}}$\text{ }\boldsymbol{\textcolor{red}{||}}\text{ }$\textcolor{blue}{\textbf{driver}},    drivers$\text{ }\boldsymbol{\textcolor{red}{||}}\text{ }$\textcolor{blue}{\textbf{pedometer}}\\\hline$\text{ }\boldsymbol{\textcolor{red}{||}}\text{ }$\textcolor{blue}{\textbf{trackpad}}$\text{ }\boldsymbol{\textcolor{red}{||}}\text{ }$\textcolor{blue}{\textbf{calculator}} $\text{ }\boldsymbol{\textcolor{red}{||}}\text{ }$\textcolor{blue}{\textbf{handsfree}}$\text{ }\boldsymbol{\textcolor{red}{||}}\text{ }$\textcolor{blue}{\textbf{autofocus}}$\text{ }\boldsymbol{\textcolor{red}{||}}\text{ }$\textcolor{blue}{\textbf{otg}}$\text{ }\boldsymbol{\textcolor{red}{||}}\text{ }$\textcolor{blue}{\textbf{troubleshooting}},    troubleshoot$\text{ }\boldsymbol{\textcolor{red}{||}}\text{ }$\textcolor{blue}{\textbf{airplane}}$\text{ }\boldsymbol{\textcolor{red}{||}}\text{ }$\textcolor{blue}{\textbf{mute}}$\text{ }\boldsymbol{\textcolor{red}{||}}\text{ }$\textcolor{blue}{\textbf{syncs}}$\text{ }\boldsymbol{\textcolor{red}{||}}\text{ }$\textcolor{blue}{\textbf{multitask}}$\text{ }\boldsymbol{\textcolor{red}{||}}\text{ }$\textcolor{blue}{\textbf{backlight}}\\\hline$\text{ }\boldsymbol{\textcolor{red}{||}}\text{ }$\textcolor{blue}{\textbf{permissions}} $\text{ }\boldsymbol{\textcolor{red}{||}}\text{ }$\textcolor{blue}{\textbf{reminders}}$\text{ }\boldsymbol{\textcolor{red}{||}}\text{ }$\textcolor{blue}{\textbf{echo}}  $\text{ }\boldsymbol{\textcolor{red}{||}}\text{ }$\textcolor{blue}{\textbf{trackball}}$\text{ }\boldsymbol{\textcolor{red}{||}}\text{ }$\textcolor{blue}{\textbf{panorama}}$\text{ }\boldsymbol{\textcolor{red}{||}}\text{ }$\textcolor{blue}{\textbf{speech}}$\text{ }\boldsymbol{\textcolor{red}{||}}\text{ }$\textcolor{blue}{\textbf{lockscreen}}$\text{ }\boldsymbol{\textcolor{red}{||}}\text{ }$\textcolor{blue}{\textbf{vga}}$\text{ }\boldsymbol{\textcolor{red}{||}}\text{ }$\\\hline

		\end{tabular}
	\end{center}
\end{table*}

\begin{figure*}
	\begin{center}
		\includegraphics[width=0.95\linewidth]{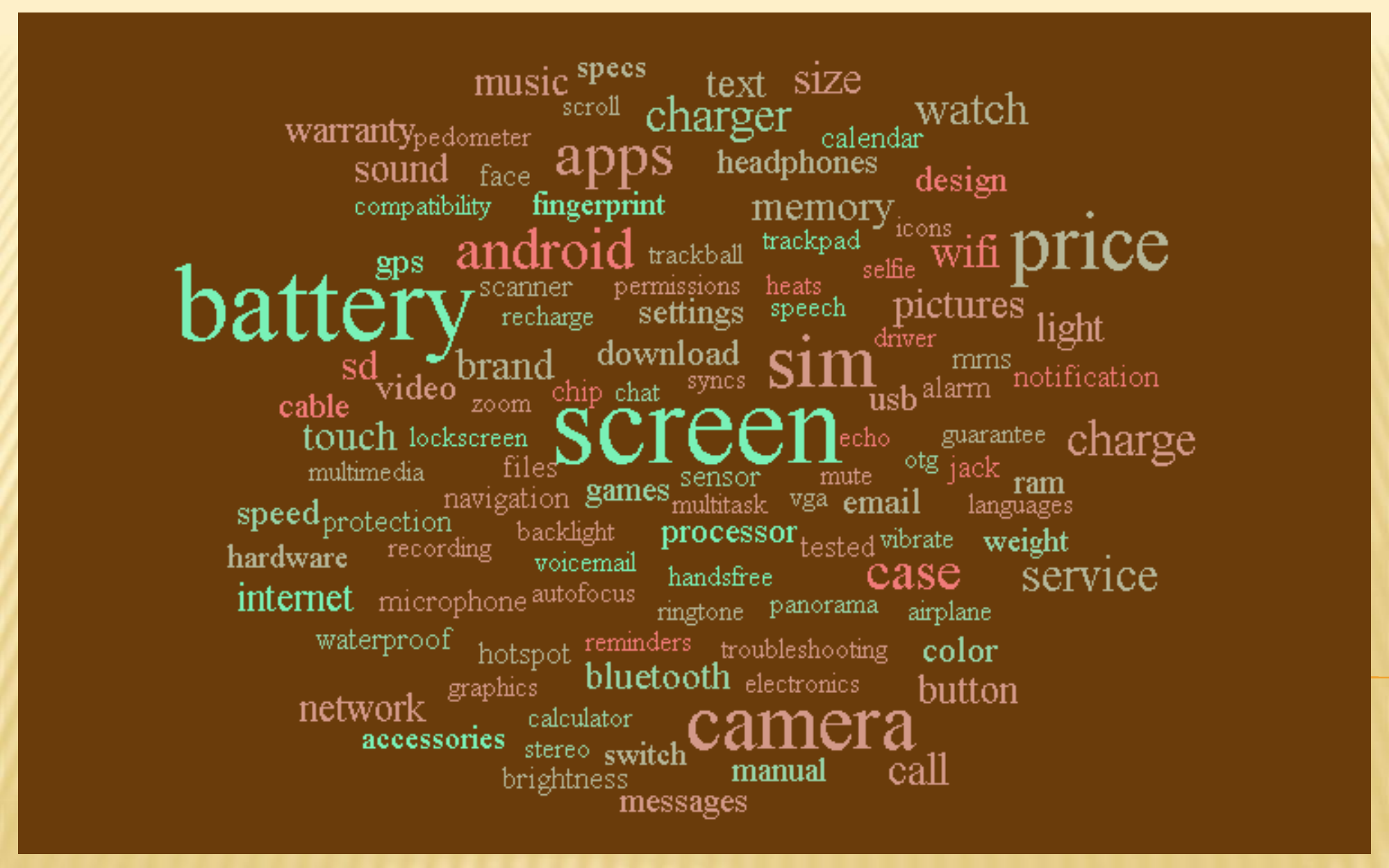}
		\caption{Word Cloud for our features (except `phone' feature).}\label{fig:wc}
	\end{center}
\end{figure*}
\section{Experiments}
In this section, we give details of experiments conducted using the proposed methodology. First, we discuss the dataset used. Then, we discuss the features chosen from the word frequency table of the dataset. At last, we discuss our analysis of the feature-level ratings obtained using the proposed method.   
\subsection{Dataset}
We apply the proposed method on a dataset named “Amazon Reviews: Unlocked Mobile Phones” \footnote{https://www.kaggle.com/PromptCloudHQ/amazon-reviews-unlocked-mobile-phones/data}, a dataset extracted by PromptCloud from the Amazon website. It consists of reviews, product-level ratings, and review votes for a total of 4418 mobile phones. There are a total of 413841 reviews present in the dataset, along with the votes obtained by them. So, there are enough reviews to carry out our sentiment analysis and obtain general insights.

\begin{table}
	\caption{Our sample results for a phone named \emph{Nokia C6}.}\label{tab:sample}
	\begin{center}
		\begin{tabular}{|l|l||l|l|}
			\hline
			Feature         & Rating & Feature    & Rating\\\hline\hline
			alarm       & 1.818 & email       & 3.107 \\ speed      & 4.977 & brightness      & 3     \\
			phone       & 3.524 & button      & 3.757 \\ chip       & 3.349 & weight          & 3.652 \\
			music       & 3.916 & sensor      & 3     \\ camera     & 4.032 & face            & 4.538 \\
			electronics & 1     & service     & 3.123 \\ calendar   & 3.305 & files           & 3.022 \\
			navigation  & 4.111 & call        & 3.785 \\ sim        & 3.255 & troubleshooting & 5     \\
			multimedia  & 3     & cable       & 3.848 \\ tested     & 3.5   & zoom            & 4.788 \\
			vibrate     & 2     & multitask   & 5     \\ network    & 3.464 & settings        & 3.169 \\
			sound       & 3.512 & gps         & 4.879 \\ internet   & 3.985 & hardware        & 2.2   \\
			charge      & 3.216 & touch       & 3.66  \\ languages  & 3     & stereo          & 3     \\
			case        & 3.454 & echo        & 3.714 \\ jack       & 3     & headphones      & 3.071 \\
			price       & 3.13  & memory      & 3.227 \\ calculator & 2     & hotspot         & 3     \\
			text        & 2.426 & wifi        & 4.702 \\ heats      & 4     & scroll          & 3.705 \\
			light       & 4.075 & specs       & 5     \\ brand      & 3.15  & usb             & 3.286 \\
			icons       & 4.667 & manual      & 3.614 \\ size       & 4.776 & warranty        & 2.773 \\
			switch      & 3.957 & messages    & 3.61  \\ apps       & 3.738 & download        & 3.787 \\
			protection  & 3.769 & battery     & 3.754 \\ android    & 3.514 & ringtone        & 3.333 \\
			chat        & 3.917 & pictures    & 3.836 \\ watch      & 2     & bluetooth       & 3.993 \\
			fingerprint & 4.206 & accessories & 2     \\ design     & 4.6   & games           & 4.016 \\
			sd          & 2.739 & screen      & 3.516 \\ mms        & 5     &  video               & 3.24       \\
			charger     & 3.298 & color       & 3.647 \\\hline
		\end{tabular}
	\end{center}
\end{table}

\begin{table}[t]
	\caption{Top few phones after ranking the phones according to the number of features in which they are the best, as per our ratings.}\label{tab:prank}
	\begin{tabular}{|l|l|}
		\hline
		Phone & No. of \\
		& features\\\hline
		\hline
		Nokia N9 - Black                                   & 17 \\ \hline
		ASUS ZenFone 3 ZE552KL (SHIMMER GOLD)              & 15 \\ \hline
		Asus ZenFone 3 ZE552KL Moonlight White             & 14 \\ \hline
		5.5" JUNING Blue                                   & 14 \\ \hline
		JUNING 7-Inch - Black                              & 14 \\ \hline
		5.5" JUNING Black                                  & 14 \\ \hline
		5.5" JUNING White                                  & 14 \\ \hline
		Asus ZenFone 3 ZE552KL Sapphire Black              & 14 \\ \hline
		LG V10 H962 64GB Ocean Blue, 5.7"                  & 12 \\ \hline
		SKY Devices Platinum Series 5.0W - Silver          & 12 \\ \hline
		LG V10 H962 64GB Black, 5.7"                       & 12 \\ \hline
		Sony Ericsson W995a Walkman (Progressive Black)    & 12 \\ \hline
		BLU Life View L110X 5.7-Inch (Blue)                & 12 \\ \hline
		THL 5000 5" FHD IPS MTK6592T (Black)               & 12 \\ \hline
		5.5" MTK6580 JUNING GSM/3G Black                   & 12 \\ \hline
		LG V10 H962 64GB 5.7-Inch (Brown Beige)            & 12 \\ \hline
		SKY Devices Platinum Series 5.0W - White           & 12 \\ \hline
		LG V10 H962 64GB 5.7-Inch (Opal Blue) (Blue White) & 12 \\ \hline
		LG V10 H962 64GB White, 5.7"                       & 12 \\ \hline
		Huawei P9 Lite VNS-L22 5.2-Inch (BLACK)            & 11 \\ \hline
		LG Electronics G3 Stylus D690 (Black Titanium)     & 11 \\ \hline
		Huawei Mate 8 32GB 6-Inch (Silver)                 & 11 \\ \hline
		Huawei Mate 8 NXT-L29 32GB 6-Inch (Space Gray)     & 11 \\ \hline
		LG VX8500 Chocolate Phone (Verizon Wireless)       & 11 \\ \hline
		ASUS Zenfone 6 A600CG 6-inches White               & 11 \\ \hline
		FuturetechÂ® A6 4.5 Inch Mtk6582 (Yellow)          & 11 \\ \hline
		Nokia C7 Unlocked Quadband Smartphone              & 11 \\ \hline
		Huawei P9 Lite VNS-L22 5.2-Inch (WHITE)            & 11 \\ \hline
		ASUS ZENFONE 6 A601CG 6" (Black)                   & 11 \\ \hline
		Samsung Galaxy S2 PLUS i9150P blue-grey            & 11 \\ \hline
		FuturetechÂ® A6 4.5 Inch Mtk6582 (black)           & 11 \\ \hline
		Nokia Asha 302                               & 10 \\ \hline
		ZTE Axon Pro Phthalo Blue                    & 10 \\ \hline
		Smart watch, GEEKERA Watch Phone( Black )    & 10 \\ \hline
		Yezz Andy 5E - (White )                      & 10 \\ \hline
		Nokia Lumia 1520 - Black                     & 10 \\ \hline
		Nokia N82 (Silver)                           & 10 \\ \hline
		Huawei P9 Lite VNS-L22 (GOLD)                & 10 \\ \hline
		Blackberry Torch 9800 - Black                & 10 \\ \hline
		HTC One Mini 2 16GB - Silver                 & 10 \\ \hline
		ZTE Spro 2 Smart Projector (Silver)          & 10 \\ \hline
		Nokia N79 (Silver)                           & 10 \\ \hline
		Cubot X15 5.5'' Inches                       & 10 \\ \hline
		Samsung Evergreen A667 - Black               & 10 \\ \hline
		Straight Talk Phone X2                       & 10 \\ \hline
		ZTE Axon Pro                                 & 10 \\ \hline
		Nokia Lumia 1520 - Red                       & 10 \\ \hline
		Honor 8 Dual Camera - Pearl White            & 10 \\ \hline
		ZTE Axon Pro, A1P133, 32 GB, Chromium Silver & 10 \\ \hline
		ZTE Axon Pro, 64 GB, Chromium Silver         & 10 \\ \hline
		Unnecto Air 5.5 (Gray)                       & 10 \\ \hline
		LG G3 Stylus 3G D690 (White)                 & 10 \\ \hline
		Nokia X3                                     & 10 \\ \hline
		OnePlus White 5.5 inch                       & 10 \\ \hline
	\end{tabular}
\end{table}
\begin{table*}[t]
	\caption{Number of phones with different integer ratings for each feature}\label{tab:numf}
	\begin{center}
		\begin{tabular}{|l|c|c|c|c|c||l|c|c|c|c|c|}
			\hline
			Feature     & 1-star & 2-star & 3-star& 4-star & 5-star & Feature         & 1-star & 2-star & 3-star& 4-star & 5-star\\
			\hline
			phone       & 12 & 88  & 1897 & 1925 & 219 & protection      & 30 & 133 & 416 & 285 & 108 \\
			screen      & 28 & 213 & 1161 & 1162 & 333 & notification    & 21 & 100 & 214 & 143 & 43  \\
			battery     & 57 & 305 & 1346 & 813  & 267 & brightness      & 5  & 8   & 82  & 199 & 245 \\
			camera      & 38 & 219 & 706  & 1049 & 466 & jack            & 23 & 113 & 247 & 116 & 50  \\
			price       & 39 & 198 & 1101 & 1575 & 378 & chip            & 17 & 105 & 287 & 165 & 71  \\
			sim         & 36 & 260 & 1336 & 811  & 182 & tested          & 18 & 132 & 515 & 328 & 134 \\
			apps        & 35 & 203 & 1104 & 797  & 216 & scanner         & 5  & 25  & 113 & 61  & 54  \\
			android     & 32 & 237 & 1097 & 803  & 219 & files           & 8  & 69  & 211 & 143 & 104 \\
			case        & 30 & 242 & 1128 & 853  & 242 & microphone      & 34 & 160 & 209 & 83  & 54  \\
			charge      & 52 & 319 & 1469 & 489  & 128 & navigation      & 15 & 65  & 198 & 154 & 112 \\
			charger     & 63 & 324 & 1260 & 458  & 119 & waterproof      & 8  & 35  & 125 & 97  & 48  \\
			service     & 51 & 227 & 848  & 725  & 331 & calendar        & 11 & 34  & 172 & 100 & 61  \\
			watch       & 22 & 98  & 311  & 282  & 171 & mms             & 4  & 27  & 55  & 51  & 21  \\
			size        & 10 & 52  & 384  & 802  & 526 & alarm           & 48 & 248 & 122 & 39  & 27  \\
			call        & 39 & 244 & 1194 & 730  & 197 & hotspot         & 12 & 20  & 144 & 83  & 41  \\
			wifi        & 36 & 183 & 573  & 320  & 137 & graphics        & 10 & 21  & 88  & 126 & 197 \\
			brand       & 29 & 115 & 705  & 598  & 314 & icons           & 16 & 116 & 251 & 163 & 52  \\
			memory      & 42 & 240 & 1124 & 639  & 214 & selfie          & 11 & 26  & 90  & 135 & 125 \\
			pictures    & 28 & 202 & 775  & 948  & 394 & recharge        & 24 & 77  & 266 & 112 & 33  \\
			touch       & 55 & 235 & 614  & 529  & 209 & electronics     & 16 & 51  & 98  & 111 & 62  \\
			text        & 39 & 204 & 842  & 468  & 192 & vibrate         & 18 & 115 & 287 & 118 & 29  \\
			sound       & 62 & 300 & 585  & 787  & 322 & recording       & 28 & 33  & 141 & 122 & 85  \\
			sd          & 25 & 119 & 525  & 384  & 160 & zoom            & 19 & 81  & 163 & 130 & 78  \\
			network     & 40 & 185 & 845  & 484  & 141 & chat            & 15 & 24  & 107 & 133 & 82  \\
			button      & 43 & 269 & 1109 & 595  & 148 & ringtone        & 16 & 60  & 161 & 81  & 57  \\
			music       & 23 & 155 & 573  & 574  & 270 & heats           & 25 & 136 & 225 & 104 & 53  \\
			internet    & 39 & 213 & 943  & 579  & 207 & voicemail       & 5  & 68  & 133 & 53  & 37  \\
			light       & 36 & 117 & 607  & 630  & 317 & stereo          & 4  & 24  & 67  & 66  & 55  \\
			warranty    & 47 & 277 & 701  & 221  & 48  & scroll          & 22 & 91  & 253 & 170 & 93  \\
			bluetooth   & 18 & 146 & 485  & 349  & 138 & guarantee       & 13 & 33  & 86  & 212 & 78  \\
			video       & 18 & 139 & 514  & 620  & 318 & languages       & 5  & 58  & 104 & 62  & 31  \\
			settings    & 15 & 195 & 749  & 554  & 168 & multimedia      & 1  & 22  & 49  & 43  & 42  \\
			color       & 32 & 104 & 305  & 565  & 509 & compatibility   & 11 & 43  & 89  & 81  & 37  \\
			design      & 13 & 81  & 282  & 813  & 632 & driver          & 13 & 26  & 124 & 74  & 31  \\
			download    & 22 & 155 & 635  & 457  & 165 & pedometer       & 0  & 0   & 25  & 17  & 4   \\
			usb         & 26 & 95  & 251  & 114  & 37  & trackpad        & 4  & 10  & 13  & 7   & 3   \\
			email       & 39 & 140 & 654  & 359  & 145 & calculator      & 2  & 24  & 57  & 42  & 32  \\
			speed       & 27 & 90  & 412  & 458  & 344 & handsfree       & 7  & 18  & 52  & 60  & 18  \\
			headphones  & 62 & 255 & 629  & 393  & 144 & autofocus       & 8  & 36  & 38  & 30  & 28  \\
			gps         & 29 & 31  & 104  & 74   & 46  & otg             & 0  & 6   & 4   & 6   & 1   \\
			games       & 22 & 89  & 330  & 434  & 256 & troubleshooting & 11 & 30  & 101 & 50  & 14  \\
			ram         & 7  & 27  & 220  & 151  & 88  & airplane        & 2  & 23  & 59  & 37  & 6   \\
			messages    & 27 & 198 & 688  & 299  & 107 & mute            & 6  & 17  & 68  & 28  & 17  \\
			cable       & 53 & 247 & 635  & 321  & 84  & syncs           & 3  & 21  & 25  & 32  & 19  \\
			manual      & 38 & 267 & 656  & 299  & 122 & multitask       & 0  & 12  & 37  & 18  & 45  \\
			processor   & 22 & 73  & 333  & 344  & 180 & backlight       & 6  & 13  & 35  & 19  & 17  \\
			specs       & 17 & 69  & 344  & 342  & 231 & permissions     & 0  & 5   & 49  & 10  & 11  \\
			hardware    & 23 & 160 & 289  & 221  & 163 & reminders       & 0  & 18  & 29  & 21  & 10  \\
			fingerprint & 35 & 134 & 437  & 320  & 138 & echo            & 14 & 22  & 54  & 15  & 4   \\
			switch      & 45 & 152 & 610  & 346  & 147 & trackball       & 1  & 3   & 9   & 15  & 0   \\
			accessories & 17 & 92  & 410  & 330  & 186 & panorama        & 0  & 1   & 25  & 20  & 23  \\
			weight      & 15 & 61  & 355  & 365  & 327 & speech          & 4  & 7   & 22  & 24  & 15  \\
			sensor      & 20 & 71  & 211  & 171  & 77  & lockscreen      & 1  & 5   & 7   & 11  & 13  \\
			face        & 23 & 102 & 206  & 181  & 148 & vga             & 0  & 0   & 5   & 2   & 7   \\\hline   
		\end{tabular}
	\end{center}
\end{table*}
\subsection{Features}
In Table~\ref{tab:words}, we list all the words which we select as feature words. As discussed earlier, they have been extracted while observing the word frequency table of the entire dataset. The feature words which are related are separated from others using \textcolor{red}{$||$}. Many of the words are just plurals of the already existing words. The related feature words are led by a feature keyword (represented in blue color), which is most prominent in the dataset amongst all the related feature words. Note that some feature words like sd, vga, otg, etc., are not common and likely to be misunderstood as incorrect words by auto-corrector. As a result, when autocorrection is applied, they may get changed to some so-called correct words. To avoid this from happening, in Eqn.\eqref{ceq}, we check the original token's presence as well in our feature words dataset, not just the corrected token. 

Note that we consider the entire phone as one of the features as well, and this is the most discussed feature, as expected. Such consideration helps us evaluate the performance of the proposed method by comparing the feature level ratings of our phone feature with the corresponding product-level ratings already available in the dataset. We also give word cloud for actual feature keywords (i.e., except phone) in Fig.~\ref{fig:wc}. Larger the keyword, most discussed it is in the reviews. It is clear from the word cloud that battery, screen, price, camera, sim, and apps are some of the most discussed features in phones.

\begin{table*}
	\setlength{\tabcolsep}{1pt} 
	\renewcommand{\arraystretch}{1}
	\caption{The best phones we recommend for different features.}\label{tab:best_phone}
	\begin{center}
		\begin{tabular}{|l|l||l|l|}
			\hline
			Features    & Best Phones                                & Features        & Best Phones\\\hline
			phone       & ASUS ZenFone 3 ZE552KL 5.5-inch (SHIMMER GOLD)       & protection      & SKY Devices Platinum Series 5.0W - Silver            \\
			screen      & LG VX8500 Chocolate Phone (Verizon Wireless)         & notification    & Blackberry Z10 16GB - Black                          \\
			battery     & JUNING 7-Inch - Black                                & brightness      & ASUS Zenfone 6 A600CG White                          \\
			camera      & Huawei P9 Lite VNS-L22 5.2-Inch (BLACK)              & jack            & Nokia N9 16 GB MeeGo OS - Black                      \\
			price       & THL 5000 5" FHD IPS MTK6592T (Black)                 & chip            & Nokia N9 16 GB MeeGo OS - Black                      \\
			sim         & Huawei P9 Lite VNS-L22 5.2-Inch (BLACK)              & tested          & Nokia N9 16 GB MeeGo OS - Black                      \\
			apps        & ASUS ZenFone 3 ZE552KL 5.5-inch (SHIMMER GOLD)       & scanner         & ASUS ZenFone 3 ZE552KL 5.5-inch (SHIMMER GOLD)       \\
			android     & LG V10 H962 64GB Ocean Blue, Dual Sim, 5.7"          & files           & ASUS ZenFone 3 ZE552KL 5.5-inch (SHIMMER GOLD)       \\
			case        & Nokia N9 16 GB MeeGo OS - Black                      & microphone      & 5.5" JUNING Blue                                     \\
			charge      & Nokia N9 16 GB MeeGo OS - Black                      & navigation      & Nokia N9 16 GB MeeGo OS - Black                      \\
			charger     & 5.5" JUNING Blue                                     & waterproof      & Sony Xperia Z1 (C6902) - Black                       \\
			service     & THL 5000 5" FHD IPS MTK6592T (Black)                 & calendar        & 5.5" JUNING Blue                                     \\
			watch       & 5.5" JUNING Blue                                     & mms             & New Genuine Nokia X3-00 Unlocked GSM X3              \\
			size        & 5.5" JUNING Blue                                     & alarm           & BLU Life View L110X 5.7-Inch (Blue)                  \\
			call        & ASUS ZenFone 3 ZE552KL 5.5-inch (SHIMMER GOLD)       & hotspot         & 5.5" JUNING Blue                                     \\
			wifi        & THL 5000 5" FHD IPS MTK6592T (Black)                 & graphics        & JUNING 7-Inch - Black                                \\
			brand       & JUNING 7-Inch - Black                                & icons           & Sony Ericsson W995a Walkman (Progressive Black)      \\
			memory      & ASUS ZenFone 3 ZE552KL 5.5-inch (SHIMMER GOLD)       & selfie          & 5.5" JUNING Blue                                     \\
			pictures    & ASUS ZenFone 3 ZE552KL 5.5-inch (SHIMMER GOLD)       & recharge        & BLU Life View L110X 5.7-Inch (Blue)                  \\
			touch       & 5.5" JUNING Blue                                     & electronics     & Samsung S7 Galaxy (SM-G930UZKAXAA)                   \\
			text        & ASUS ZenFone 3 ZE552KL 5.5-inch (SHIMMER GOLD)       & vibrate         & Sony Ericsson W995a Walkman (Progressive Black)      \\
			sound       & SKY Devices Platinum Series 5.0W - Silver            & recording       & ZTE Axon Pro, 64 GB Phthalo Blue                     \\
			sd          & Nokia N9 16 GB MeeGo OS - Black                      & zoom            & Nokia N9 16 GB MeeGo OS - Black                      \\
			network     & ASUS ZenFone 3 ZE552KL 5.5-inch (SHIMMER GOLD)       & chat            & BLU Life View L110X 5.7-Inch (Blue)                  \\
			button      & ASUS ZenFone 3 ZE552KL 5.5-inch (SHIMMER GOLD)       & ringtone        & Samsung Evergreen A667 - Black                       \\
			music       & Nokia N9 16 GB MeeGo OS - Black                      & heats           & THL 5000 5" FHD IPS MTK6592T (Black)                 \\
			internet    & Nokia N9 16 GB MeeGo OS - Black                      & voicemail       & ZTE Axon Pro, 64 GB Phthalo Blue                     \\
			light       & JUNING 7-Inch - Black                                & stereo          & BLU Life View L110X 5.7-Inch (Blue)                  \\
			warranty    & Samsung Galaxy Note 5, WhiteÂ  32GB (AT\&T)          & scroll          & LG V10 H962 64GB Ocean Blue, Dual Sim, 5.7"          \\
			bluetooth   & LG VX8500 Chocolate Phone (Verizon Wireless)         & guarantee       & Smart watch, GEEKERA Bluetooth Watch Phone ( Black ) \\
			video       & JUNING 7-Inch - Black                                & languages       & Sony Ericsson W995a Walkman (Progressive Black)      \\
			settings    & Asus ZenFone 3 ZE552KL 64GB Sapphire Black, 5.5-inch & multimedia      & LG V10 H901 64GB T-Mobile- Space Black               \\
			color       & Nokia N9 16 GB MeeGo OS - Black                      & compatibility   & LG Electronics G3 Stylus D690 (Black Titanium)       \\
			design      & Nokia N9 16 GB MeeGo OS - Black                      & driver          & BLU Life View L110X 5.7-Inch (Blue)                  \\
			download    & ASUS ZenFone 3 ZE552KL 5.5-inch (SHIMMER GOLD)       & pedometer       & Sony Ericsson W995a Walkman (Progressive Black)      \\
			usb         & SKY Devices Platinum Series 5.0W - Silver            & trackpad        & BlackBerry Passport (SQW100-1)Black                  \\
			email       & 5.5" JUNING Blue                                     & calculator      & 5.5" JUNING Blue                                     \\
			speed       & Nokia N9 16 GB MeeGo OS - Black                      & handsfree       & LG G4 5.5-Inch (Black Leather)                       \\
			headphones  & ASUS ZenFone 3 ZE552KL 5.5-inch (SHIMMER GOLD)       & autofocus       & BLU ENERGY X - Gold                                  \\
			gps         & Sony Ericsson W995a Walkman (Progressive Black)      & otg             & BLU ENERGY X - Gold                                  \\
			games       & Nokia N9 16 GB MeeGo OS - Black                      & troubleshooting & Sony Xperia sola MT27i-BLK (Black)                   \\
			ram         & ASUS Zenfone 6 A600CG White                          & airplane        & Kyocera Hydro C5170                                  \\
			messages    & Blackberry Torch 9800 - Black                        & mute            & HTC One M7 - Black                                   \\
			cable       & Nokia N9 16 GB MeeGo OS - Black                      & syncs           & Nokia Lumia 925 RM-893 - Black/Dark Grey             \\
			manual      & ASUS ZenFone 3 ZE552KL 5.5-inch (SHIMMER GOLD)       & multitask       & ZTE Axon Pro, 64 GB Phthalo Blue                     \\
			processor   & ASUS ZenFone 3 ZE552KL 5.5-inch (SHIMMER GOLD)       & backlight       & ZTE Axon Pro, 64 GB Phthalo Blue                     \\
			specs       & ASUS ZenFone 3 ZE552KL 5.5-inch (SHIMMER GOLD)       & permissions     & POSH Titan HD E500a - 5.0" HD, (Yellow)              \\
			hardware    & ASUS ZenFone 3 ZE552KL 5.5-inch (SHIMMER GOLD)       & reminders       & BLU Life View L110X 5.7-Inch (Blue)                  \\
			fingerprint & LG V10 H962 64GB Ocean Blue, Dual Sim, 5.7"          & echo            & Motorola Moto E (1st Generation) - Black - 4 GB      \\
			switch      & Nokia N9 16 GB MeeGo OS - Black                      & trackball       & Blackberry Gemini 8520 White                         \\
			accessories & ZTE Axon Pro, 64 GB Phthalo Blue                     & panorama        & ZTE Axon Pro, 64 GB Phthalo Blue                     \\
			weight      & 5.5" JUNING Blue                                     & speech          & Verizon LG Ally VS740                                \\
			sensor      & Nokia N9 16 GB MeeGo OS - Black                      & lockscreen      & LG G3 D855 Black                                     \\
			face        & LG V10 H962 64GB Ocean Blue, Dual Sim, 5.7"          & vga             & BLU Win Jr Windows White  
			\\\hline      
		\end{tabular}
	\end{center}
\end{table*}

\subsection{Ratings}
We give sample feature-level ratings of our proposed method in Table~\ref{tab:sample}; it is for Nokia C6. It can be noted how the proposed method can rate a phone on a vast number of features, and this can certainly help consumers \cite{glenski2017consumers} in making their decisions on buying their mobile phones. Since we can get feature-level ratings for all the phones, to summarize our results, we report the number of phones close to different integer ratings (e.g., 2-star) for each feature in Table ~\ref{tab:numf}. We obtain this by rounding off the ratings calculated. It is clear from the Table that most of the phones get 3-star ratings, i.e., average, across our features. Note that not all the phones will have reviews mentioning every single feature we have selected. For example, the vga feature (at the end of the Table) is mentioned in the reviews of only 14 phones. So, only 14 phones can have ratings for the vga feature. In Table~\ref{tab:prank}, we rank all our phones based on the number of features in which they are the best based on the reviews. We report only those phones which are best for at least ten features. Based on these rankings, we also report the best phone for which the required feature has been rated in Table~\ref{tab:best_phone} so that we can recommend a phone for a given feature. For example, both “Nokia N9 - Black” and “JUNING 7-Inch - Black” have ratings of 5 for music, and we recommend “Nokia N9 - Black” since it’s best in more number of features. So, the higher the ranking, the better is the chance for the recommendation.      

\begin{table}
	\caption{Different error metric values while comparing the predicted ratings of our `phone' feature with the actual ratings given by the customers}\label{tab:err}
	\begin{center}
		\begin{tabular}{|l||l|l|l|}
			\hline
			Error Metrics & MSE   & RMSE  & MAE   \\ \hline
			Values        & 0.545 & 0.738 & 0.555 \\ \hline
		\end{tabular}
	\end{center}
\end{table}
\begin{table}
	\caption{Confusion Matrix while comparing the predicted ratings for our `phone' feature with the actual integer ratings of phones}\label{tab:cm}
	\begin{tabular}{|c|l|l|l|l|l|l|}
		\hline
		\multicolumn{1}{|l|}{}  & \multicolumn{6}{c|}{Predicted}                      \\ \hline
		\multirow{6}{*}{Actual} &        & 1-star & 2-star & 3-star & 4-star & 5-star \\ \cline{2-7} 
		& 1-star & 6      & 23     & 73     & 11     & 3      \\ \cline{2-7} 
		& 2-star & 2      & 23     & 274    & 59     & 3      \\ \cline{2-7} 
		& 3-star & 2      & 30     & 841    & 376    & 30     \\ \cline{2-7} 
		& 4-star & 2      & 9     & 649    & 1184   & 72     \\ \cline{2-7} 
		& 5-star & 0      & 3      & 60     & 295    & 111    \\ \hline
	\end{tabular}
\end{table}

\subsection{Evaluation}
Although there are no ground-truth feature-level ratings available to evaluate our method, we have the purposely selected “phone” features that can be evaluated. We can compare our results on the phone named feature with the phone’s overall ratings already available in the dataset. The ratings given by individual customers are weighted averaged (weighted by the review votes) and considered as ground-truth ratings for our phone named feature. We report different error metric values in the Table~\ref{tab:err} while comparing with such ground truth ratings. It is impressive that, on average, our rating differs from the actual ratings by just 0.555, which is approximately just half a star, as suggested by the MAE (Mean Average Error) error metric. 

Also, we report a confusion matrix for our phone feature in Table~\ref{tab:cm}. Both our and ground-truth ratings are rounded to get such integer star ratings. So, we now have five classes (1-star to 5-star) into which a phone can be classified. With such discrete outputs now, we generate the confusion matrix. It suggests that our system predicts correct ratings for 2165 mobiles and the ratings within 1-star closeness for 3886 mobiles out of a total of 4141 mobiles. So, if we want the exact integer star rating, the accuracy of the proposed method is 52.3\%. However, if we can tolerate the error of 1-star in the integer star rating, then accuracy jumps to 93.8\%. Therefore, we can comfortably say that the proposed methodology does work well for the phone named feature. Note that the total number of mobiles here has changed from 4418. That is because there might be some phones which do not have any review with the feature words we have chosen as the words related to the phone named feature. Therefore, such phones do not receive the rating for their phone named feature to participate in this kind of evaluation, which requires at least one such review for consideration.

\section*{Conclusion}
We have developed a system to rate mobile phones in terms of 108 features based on customer reviews and review votes. We could rate 4k+ phones; this can help make personalized buying decisions and improve the products. We accomplish this by first converting the unstructured data into structured data; then, we extract the sentences comprising our feature keywords; then, we were able to provide the feature-level ratings through sentiment analysis of these sentences. We rank the phones based on the number of features they are best at, and accordingly, we were able to recommend the best phones for a feature. We tested our methodology on the “phone” named feature by considering the overall customer ratings as ground-truth ratings. The performance of our method is found to be decent. We obtain MAE of only 0.555, i.e., approximately just half a star. We get 52.3\% accuracy if exact integer ratings have to be predicted. However, if we can tolerate the 1-star integer rating error, the accuracy jumps to 93.8\%. The proposed approach is unsupervised. As an extension, we will work on improving the performance by taking a weakly-supervised or supervised approach to this problem, for which we will have to annotate the available data in terms of all our 108 features. 

\ifCLASSOPTIONcaptionsoff
  \newpage
\fi

\bibliographystyle{IEEEtran}
\bibliography{main}

\begin{IEEEbiography}[{\includegraphics[width=1in,height=1.25in,clip,keepaspectratio]{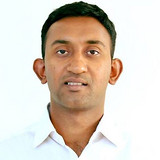}}]{Koteswar Rao Jerripothula}(Member, IEEE)) received the B.Tech. degree from Indian Institute of Technology Roorkee (IIT Roorkee), India, in 2012, and the Ph.D. degree from Nanyang Technological University (NTU), Singapore, in 2017. He is currently an assistant professor with the CSE department, Indraprastha Institute of Information Technology Delhi (IIIT Delhi), India. He has also worked with Graphic Era, ADSC (Singapore), and Lenskart earlier. His major research interests include computer vision, machine learning, and natural language processing. He was the recipient of the “Top 10\% paper” award at ICIP’14. He has published in several top venues like CVPR, ECCV, TMM, and TCSVT. 

\end{IEEEbiography}

\begin{IEEEbiography}[{\includegraphics[width=1in,height=1.25in,clip,keepaspectratio]{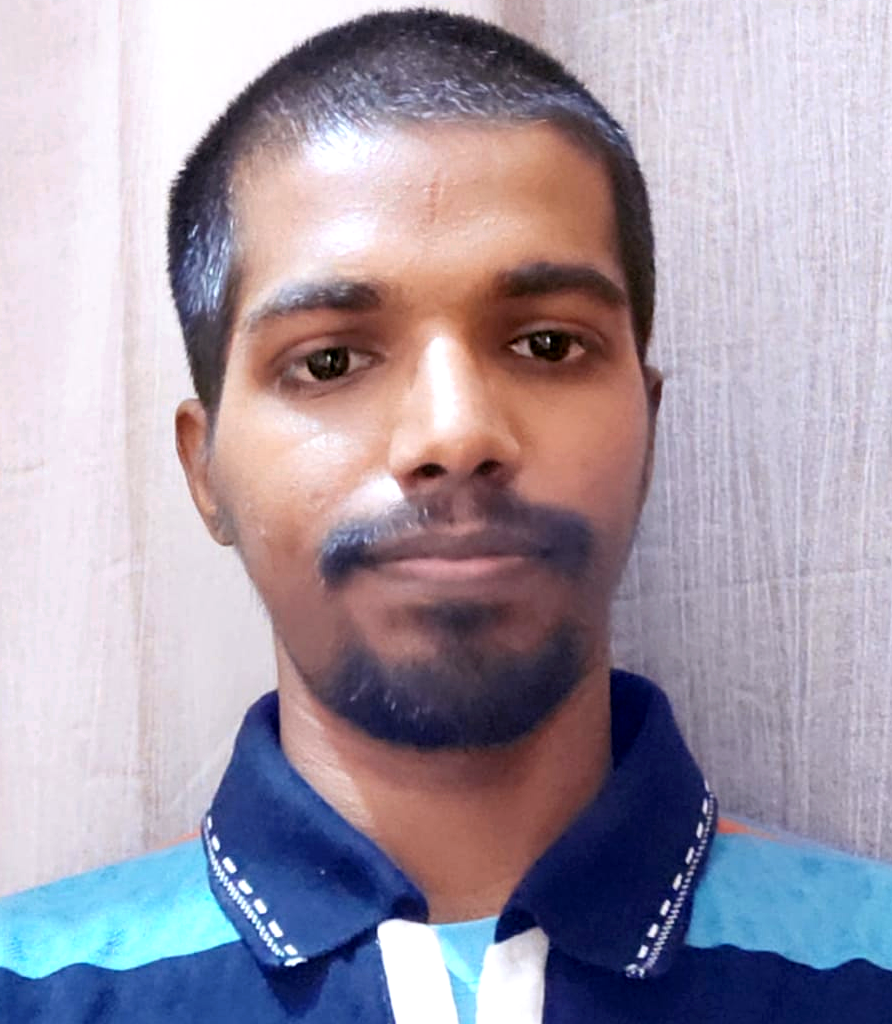}}]{Ankit Rai} received his B.Tech degree from Graphic Era, India, in 2020. He is currently working with Infosys, India. His research interests include machine learning and natural language processing. 

\end{IEEEbiography}

\begin{IEEEbiography}[{\includegraphics[width=1in,height=1.25in,clip,keepaspectratio]{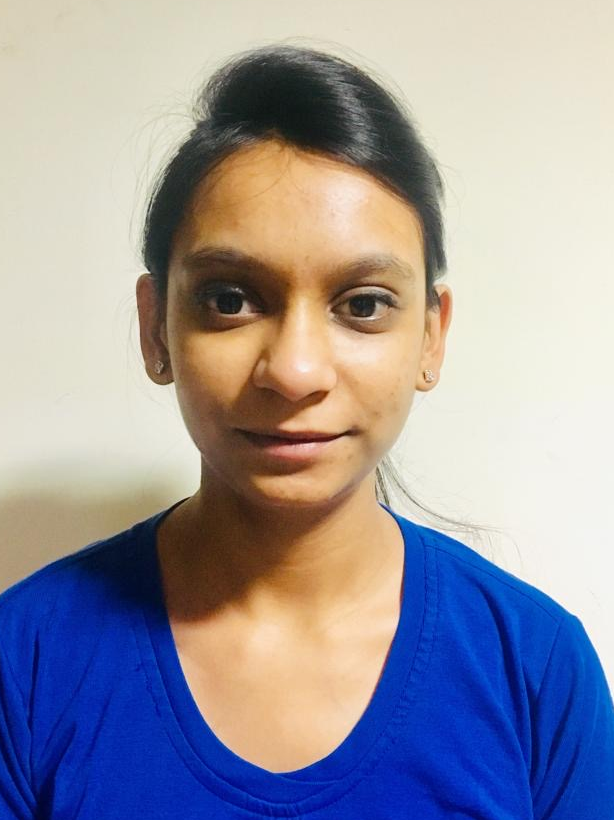}}]{Kanu Garg} received her B.Tech degree from Graphic Era, India, in 2020. She is currently working with Infosys, India. Her research interests include computer vision, machine learning and natural language processing. 

\end{IEEEbiography}

\begin{IEEEbiography}[{\includegraphics[width=1in,height=1.25in,clip,keepaspectratio]{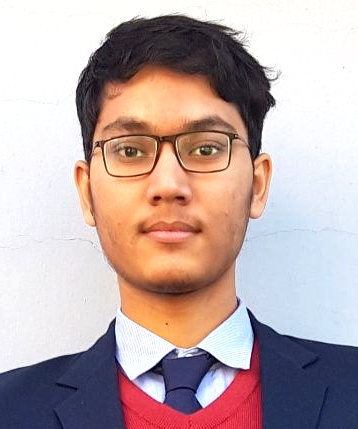}}]{Yashvardhan Singh Rautela} received his B.Tech degree from Graphic Era, India, in 2020. He is currently working with Cognizant, India. His research interests include machine learning and natural language processing.

\end{IEEEbiography}

\end{document}